\documentclass[11pt,a4paper]{article}\usepackage[hyperref]{acl2019}
\usepackage{times}
\usepackage{latexsym}

\usepackage{url}

\usepackage{epsfig}
\usepackage{graphicx}
\usepackage{amsmath}
\usepackage{amssymb}
\usepackage{latexsym}
\usepackage{graphicx}
\usepackage[colorinlistoftodos]{todonotes}
\usepackage{booktabs}
\usepackage{multirow}
\usepackage{amsmath}
\usepackage{subcaption}
\usepackage{xspace}
\usepackage{soul}
\usepackage{comment}

\graphicspath{{figures/}}
\newbox{\bigpicturebox}

\newcommand{\vifidel}{\texttt{VIFIDEL}\xspace}

\aclfinalcopy 
\def\aclpaperid{2770} 

\newcommand\blfootnote[1]{%
  \begingroup
  \renewcommand\thefootnote{}\footnote{#1}%
  \addtocounter{footnote}{-1}%
  \endgroup
}

\title{\vifidel: Evaluating the Visual Fidelity of Image Descriptions}

\author{Pranava Madhyastha$^{\Leftrightarrow}$, Josiah Wang$^{\Leftrightarrow}$ \and Lucia Specia\\
Department of Computing \\
Imperial College London, UK \\
  {\tt {\{pranava, josiah.wang, l.specia\}@imperial.ac.uk}}
}

\date{}

\begin{document}
\maketitle
\begin{abstract}
We address the task of evaluating image description generation systems. We propose a novel image-aware metric for this task: \vifidel. It estimates the faithfulness of a generated caption with respect to the content of the actual image, based on the semantic similarity between labels of objects depicted in images and words in the description. The metric is also able to take into account the relative importance of objects mentioned in human reference descriptions during evaluation. Even if these human reference descriptions are not available, \vifidel can still reliably evaluate system descriptions. The metric achieves high correlation with human judgments on two well-known datasets and is competitive with metrics that depend on human references.
\end{abstract}


\section{Introduction}
\label{sec:intro}

A popular task at the intersection of computer vision and natural language is \emph{image description generation} (IDG), i.e.\ the task of generating as output a sentence describing the visual content of a given input image. 
\blfootnote{$\Leftrightarrow$ 
  Pranava Madhyastha and Josiah Wang contributed equally  as joint first authors to this work.
  }
While a variety of methods have been proposed for this task \cite{KulkarniEtAl:2011,LiEtAl:2011,VinyalsEtAl:2015}, its evaluation is still an understudied problem. Evaluation of IDG is currently performed in two ways: (i) human judgment; (ii) automatic metrics. Human judgments evaluate either the overall quality of descriptions or specific criteria in isolation (relevance, fluency, etc.). Such methods, however, can be subjective and expensive to scale.

Automatic metrics address the scalability issue by comparing candidate descriptions against human-authored reference descriptions. These metrics conflate various criteria implicitly into a single evaluation assumption, i.e.\ a good description is one that is similar to one or more human-authored descriptions, presuming that these gold descriptions are fluent, correct and relevant to the image. Existing automatic metrics are thus useful for measuring the quality of descriptions as a whole, but this makes it difficult for the specific capabilities of IDG systems to be inspected.

\begin{figure}[t]
\centering
\begin{subfigure}{0.35\linewidth}
\includegraphics[width=\linewidth]{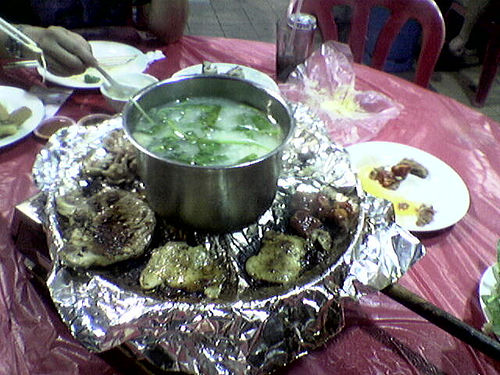}
\caption{`What is this??' 
}
\label{fig:pascalinfidel}
\end{subfigure}
\hfill
\begin{subfigure}{.35\linewidth}
\includegraphics[width=\linewidth]{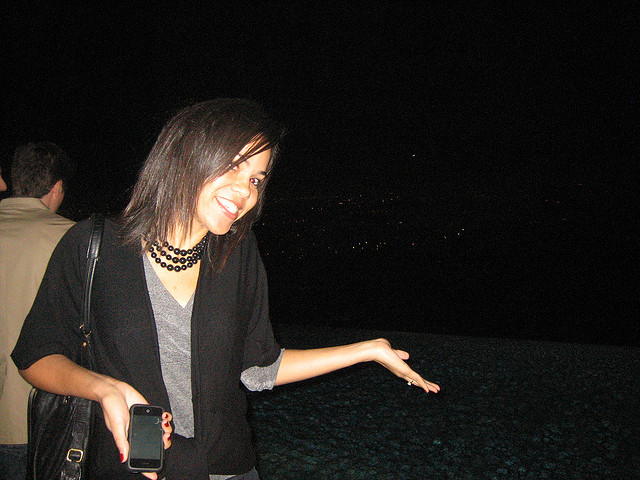}
\caption{`so what else is she supposed to do?' 
}
\label{fig:cocoinfidel}
\end{subfigure}

\caption{Examples of inadequate references taken from (a) PASCAL-50S and (b) MSCOCO.}
\label{fig:infidel}
\end{figure}

We argue that a fine-grained metric measuring specific criteria would be more useful in understanding \emph{how} an IDG system is better than another. We focus on one such criterion, \emph{visual fidelity}\footnote{Also referred to as \emph{relevance}, \emph{faithfulness} or \emph{correctness}.}. This criterion aims to measure how faithful a description is with respect to what is depicted in the image (i.e.\ systems should be rewarded for describing elements depicted in the image and penalised for describing things that are not depicted). 
For that, we propose to take \emph{image content} into account when evaluating descriptions, in contrast to previous work ($\S$\ref{sec:related}) that rely solely on words in the reference descriptions. 
Given that most datasets are crowd-sourced, reference descriptions may not always accurately describe the image (describing non-depicted objects, not mentioning relevant objects, e.g.\ Figure~\ref{fig:infidel}). For reliable evaluation, multiple references are needed. Image annotations (with objects, attributes, relations, etc.), on the other hand, are 
arguably more general and less ambiguous for evaluating visual fidelity: they require a single annotation per image, and are not affected by language or style preferences. 
To our knowledge, no existing metric for IDG has images factored explicitly into the evaluation process.

\begin{figure}[t]
\centering
\includegraphics[width=\linewidth]{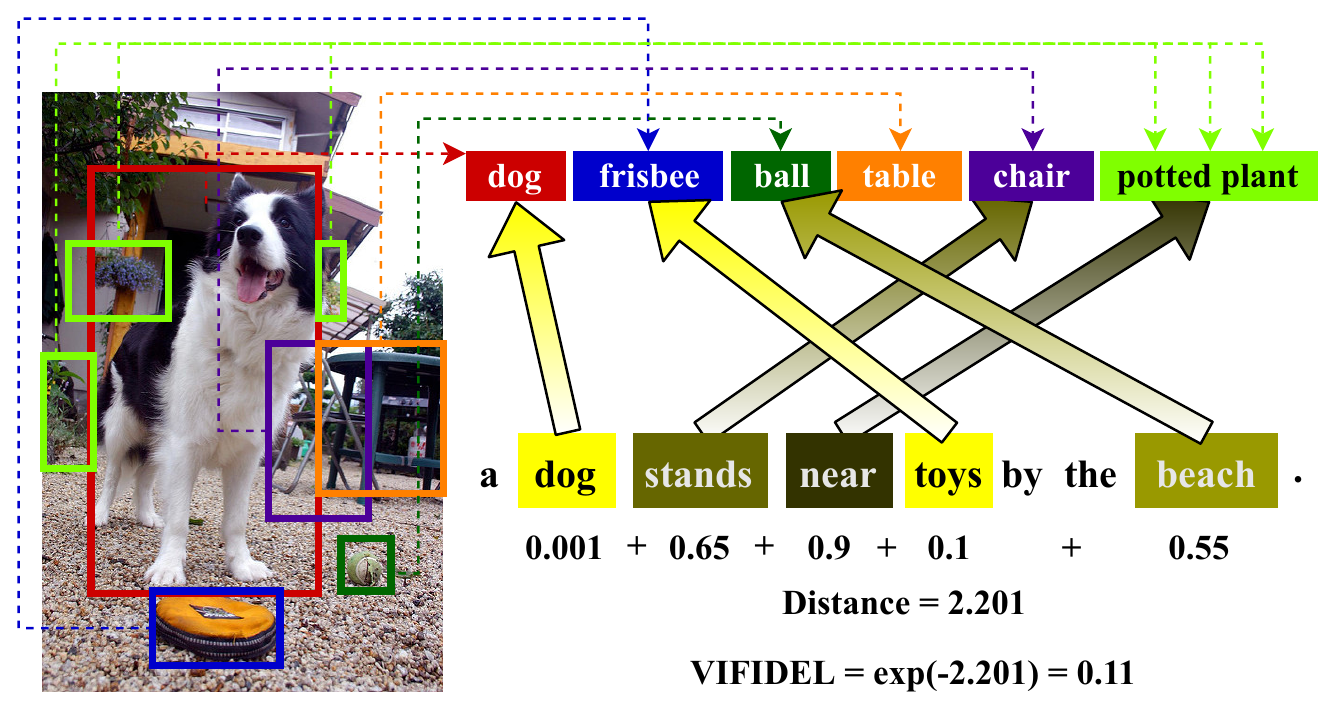}
\caption{An illustration of \vifidel as a visual fidelity metric for IDG. Bright yellow arrows indicate that the word is semantically similar to the object according to Word Mover's Distance, while darker arrows show that they are less related  (larger distance).}
\label{fig:overview}
\end{figure}

Our {\bf main contribution} is therefore an automatic  evaluation metric for IDG that measures the fidelity of image descriptions with respect to the image, using information derived from images directly as `reference' (Figure~\ref{fig:overview}). We name it \vifidel (VIsual Fidelity for Image Description EvaLuation). \vifidel can be used (i) based only on images as reference (no textual references) ($\S$\ref{sec:vifidel-images}), or (ii)  in conjunction with textual references to take into account the relevant image content people describe 
($\S$\ref{sec:vifidel-references}). In addition, \vifidel performs matching of images and text in an embeddings space, thus drawing both modalities together semantically while avoiding the pitfalls of mainstream metrics that rely on exact or approximate string matching. This is done by building on the Word Mover's Distance (WMD) metric, which measures the distance between two texts in a word embeddings space. Another contribution is the extension of WMD to allow for multiple references to be used to model object importance, i.e.\ an approach for consensus within WMD. 
We evaluate the performance of \vifidel against human judgments on two popular IDG datasets ($\S$\ref{sec:evaluation}).


\section{Background}
\label{sec:related}

Various IDG metrics have either been adapted from other fields or proposed specifically for IDG. Examples of the former include BLEU \cite{PapineniEtAl:2002} and Meteor~\cite{DenkowskiLavie:2014} from machine translation evaluation, and ROUGE~\cite{Lin:2004} (more specifically ROUGE$_L$~\cite{LinOch:2004}) from text summarisation evaluation. Metrics designed specifically for evaluating image descriptions include CIDEr \cite{VedantamEtAl:2015}, SPICE \cite{AndersonEtAl:2016} and BAST \cite{EllebrachtEtAl:2015}.

One main weakness of most metrics (BLEU, ROUGE, Meteor, CIDEr) is that they rely on exact string matching to measure the surface-level, $n$-gram overlap between candidate texts and human references. This can result in data spasity problems, especially with limited references. 
Meteor partially addresses this by matching synonyms from dictionaries and paraphrase tables, but it is constrained to the availability of such dictionaries, making it hard to scale to different languages. 

Metrics like SPICE and BAST address the issue of exact string matching by measuring similarity on a \emph{semantic} level. However, these methods rely heavily on linguistic resources such as parsers, semantic role labellers, tailored rules, etc., 
making evaluation difficult to scale or adapt to different languages and domains. Word Mover's Distance (WMD)~\cite{KusnerEtAl:2015} has also been proposed as an IDG metric \cite{KilickayaEtAl:2017}. WMD finds optimal alignments between word embeddings in candidate and reference descriptions instead of performing $n$-gram matching to address data sparseness issues. \vifidel is inspired by WMD, but goes beyond using reference texts by comparing candidate texts  against image content ($\S$\ref{sec:vifidel-images}). 

\paragraph{Using images for IDG evaluation.}
As far as we are aware, no previous work uses images for evaluating IDG. The closest related work is that by \citet{WangGaizauskas:2015}, who propose an $f$-measure-based metric to evaluate the task of selecting relevant object instances to be mentioned in a description. The metric computes the overlap between selected object instances and objects mentioned in references, averaged over multiple references. The averaging process implicitly captures consensus over which objects should be mentioned ($\S$\ref{sec:vifidel-references}), i.e.\ objects mentioned in more references should be more important than those mentioned in fewer. Their work, however, requires manual correspondence annotations between bounding box instances and object mentions in descriptions. Our proposed method leverages word embeddings to circumvent the need for exact correspondence annotations.

\paragraph{Complementarity of metrics.}
\citet{KilickayaEtAl:2017} report that the $n$-gram based metrics, the semantic graph based SPICE, and the embedding-based WMD capture complementary information, and that linearly combining Meteor, SPICE and WMD gives a better correlation score against human judgments. Similarly, \citet{LiuEtAl:2017} optimise image description generation to capture both semantic faithfulness and syntactic fluency by combining metrics with complementary properties. However, the combination weights need to be engineered towards the task.





\section{\vifidel}

In this section we describe the \vifidel metric.
It is inspired by Word Mover's Distance (WMD) ($\S$\ref{sec:wmd}), which measures the distance between two documents in a word embedding space. \vifidel however explicitly incorporates semantic information derived from images ($\S$\ref{sec:vifidel-images}), which can be used with or without reference descriptions ($\S$\ref{sec:vifidel-references}), using word embeddings as a bridge for matching content in images to content in textual descriptions. In contrast to WMD which compares pairs of documents, \vifidel also allows for multiple references to be taken into account with a consensus-based approach.

\subsection{Word Mover's Distance (WMD)} 
\label{sec:wmd}

WMD \cite{KusnerEtAl:2015} makes use of word vector relationships between word embeddings to compute the distance between two text documents. 
WMD captures 
the minimal distance required to move words from the first document to words in the second document. 

Let $X \in \mathcal{R}^{N{\times}K}$ be a matrix, with $K$-dimensional word embeddings for a vocabulary of $N$ words. Let $x_i \in \mathcal{R}^K$ be a $K$-dimensional word embedding vector for word $i$. A document $\Delta$ is represented as an $N$-dimensional normalised bag-of-words (BOW) vector, $d^\Delta = (d^\Delta_1, d^\Delta_2, ..., d^\Delta_N )$, where $d^\Delta_i$ is the normalised frequency of word $i$ occurring in document $\Delta$. Stop words are removed from documents; only content words are retained. \citet{KusnerEtAl:2015} state that stop words are generally less relevant for capturing semantic similarity between documents, especially for bag-of-words representations. 

As a measure of word-level dissimilarity, \citet{KusnerEtAl:2015} propose the word travel cost, that is the cost of moving from word $i$ to word $j$, using the Euclidean distance between the embeddings corresponding to words. More precisely, the cost is defined as: 
\begin{equation}
c(i, j) = \|x_i - x_j\|_2^p~,
\label{eq:cost}
\end{equation}
where $p$ is usually $1$ or $2$ (we set $p$=$2$).
This allows documents with many closely related words to have smaller distances than documents with very dissimilar words. 

To measure distances between two documents $\alpha$ and $\beta$, WMD defines a transport matrix $T \in \mathcal{R}^{N{\times}N}$,
where $T_{ij}$ contains information about the proportion of word $i$ in $d^\alpha$ that needs to be transported to word $j$ in $d^\beta$. Formally, WMD computes $T$ that optimises: 
\begin{equation}
\text{WMD}(d^\alpha, d^\beta) = \min_{T{\geq}0}\sum_{i,j=1}^{N}T_{ij}\,c(i,j)
\label{eq:WMD}
\end{equation}
such that:~$\sum_{j=1}^{N}{T_{ij}}$=$d_i^\alpha$ and~$\sum_{i=1}^{N}{T_{ij}}$=$d_j^\beta$,~$\forall{~}i,j$. Here, the normalised bag-of-words distribution of the documents $d^\alpha$ and $d^\beta$ contains a combined vocabulary from $d^\alpha$ and $d^\beta$ resulting in a square transport matrix $T$ of dimensionality $N{\times}N$. 
 
\citet{KusnerEtAl:2015} note that WMD 
is a special case of Earth Mover's Distance~\cite{RubnerEtAl:2000}, popular in the computer vision community, or Wasserstein's Distance~\cite{DattaEtAl:2008}, popular in the optimal transport community.

\subsection{Using objects as image information}
\label{sec:vifidel-images}

In this paper, we explore image information in the form of explicit object detections, both using \emph{gold} or \emph{predicted} object instances for a given image. Previous works \cite{YinOrdonez:2017,MadhyasthaEtAl:2018,WangEtAl:2018} have found explicit object detections to be informative for image description generation. Thus, we base our intuition on the  hypothesis that {\em a thorough and true description of the image should consist of information about objects and their interactions (frequencies, etc.) in the environment}. \vifidel has the capacity to capture these.

While objects represent one important type of semantic image information, \vifidel can potentially incorporate other semantic image information including attributes, actions, positions, scenes, relations between objects, and more fine-grained information such as colour. For this paper, we consider only the frequency of depicted object category instances as semantic information, and regard further enrichment as future work.

As mentioned, a motivation for using image information for evaluating image descriptions is that reference descriptions can be subjective, ambiguous and may or may not specify all (and only) the important elements of the image. In fact, they often focus on a subset of the image content. Using object labels can minimise these issues. In addition, as shown in Figure~\ref{fig:infidel}, references are not always actual descriptions and can be incorrect. 

Another advantage of a metric based on object-level information is the cost of collecting data: either objects are predicted (no labelling involved) or, if gold labels are to be used for more reliable results, object annotations can be gathered in more trustworthy ways using a single annotation per image. With descriptions, it has been shown that multiple descriptions per image are needed for reliable evaluation~\cite{VedantamEtAl:2015}. 

Figure~\ref{fig:overview} gives an intuitive illustration of our metric. The top row shows a set of detected object instances in the image. The bottom row shows  content words in a system description  that are semantically very similar to the detected objects -- dog, toys, etc. 
The description does not mention all detected objects (e.g. it misses {\em table}) and contains the word \emph{beach} that is not in the image. \vifidel aims to capture these discrepancies.


More specifically, \vifidel is defined as the similarity between the semantic content in image $I$ and a description $S$. It is specified by the inverse of the minimum cumulative cost required to move semantic labels (e.g.\ object categories) from image $I$ to words in the description $S$. This converts WMD from a distance measure to a similarity measure. Formally:  
\begin{equation}
\text{\vifidel}(I,S) = \exp(-\text{WMD}(d^I, d^S))
\label{eq:vifidel}
\end{equation}

\noindent where $d^S$ is the normalised bag of words representation for description $S$ ($\S$\ref{sec:wmd}), and $d^I$ is a semantic vector representation for image $I$, specifically a normalised bag of object category labels ($\S$\ref{sec:visual}). WMD is defined in Eq.~\ref{eq:WMD}.

In its basic form, \vifidel can provide information about the compatibility between an image and a description without using  reference descriptions. 
We show the performance of \vifidel in the absence of reference descriptions in $\S$\ref{sec:evaluation}.

\subsection{Modelling object importance with reference descriptions}
\label{sec:vifidel-references}

We now expand the basic version of \vifidel to use (one or more) human references when available. Human references allow for capturing the \emph{human-likeness} aspect of descriptions, that is, they capture what humans consider important to be described for a picture~\cite{BergEtAl:2012}. 

\vifidel can use human references as additional guidance to determine the importance of object content in a given image. We exploit the fact that each reference may only describe a particular subset of the image content, and we assume that important objects are mentioned more frequently across references than less important ones. Our proposal is similar to CIDEr in that we capture consensus information given a set of human references. However, we significantly differ in that (i) we use the references to explicitly model object importance, instead of directly comparing the candidates against the references; (ii) we perform word matching in a semantic space using word embeddings rather than surface forms. 

\vifidel also differs from previous approaches using WMD for image description evaluation~\cite{KilickayaEtAl:2017}, where the metric is only computed using the description and the single closest reference. One of the problems in such an approach is the biased choice of reference: in this case, the reference with the smallest WMD distance from the system description. A better reference may be available, e.g. mentioning more image content, which would lead lower system scores in an unbiased evaluation. This has been a common problem in metrics based on a single reference \cite{FomichevaSpecia:2016}. 

Another contribution in this paper is therefore the incorporation of an object importance model into the WMD framework using human references. Our approach rewards candidate descriptions that mention objects depicted in the image (i.e.\ faithful to image content), and that the objects are also mentioned frequently across all references (i.e.\ they mention important objects). 
Intuitively, the WMD cost function (Eq.~\ref{eq:cost}) is replaced with a \emph{weighted} Euclidean distance. These weights are derived from human descriptions. While the original cost function captures the faithfulness of candidate descriptions to depicted objects, the weights extend the function such that the cost is lower for words that are mentioned frequently across references (either as an exact match or a semantically similar match), and higher for those mentioned less frequently. The weights are applied to both the object labels from images and words for candidate descriptions.

Formally, let $R^I=(R^I_1,R^I_2,\ldots,R^I_M)$ be a set of $M$ human references for image $I$. The per-image penalty weight, $\rho^I_k$, for a word $k$ (an object label in image $I$ or a content word in a candidate description $S^I$) is computed as:

\begin{equation}
\rho^I_k = \frac{1}{M}\sum_{r=1}^{M}\bigg(\frac{1- \max_{t \in \{R^{I}_{r}\}} \cos(x_k, x_t) }{2}\bigg)
\label{eq:weight}
\end{equation}

\noindent where $\{R^{I}_{r}\}$ is the set of content words in the $r$th reference for image $I$, and $x_t$ the word embedding for word $t$. 
The denominator $2$ ensures that $w^I_k$ is always in the range $[0,1]$.

For each image $I$, we compute the penalty $\rho^I_k$ for each $k \in \{t | d^I_t > 0\} \cup \{t | d^{S_I}_t > 0\}$, i.e.\ the union of all labels for objects depicted in $I$ and all content words in the candidate description $S^I$ to be evaluated. Thus, $\rho^I_k$ is the effective cosine distance ($\in [0,1]$) between each word/object label $k$ and its most similar content word in each human reference, averaged over all references for the image. The averaging process implicitly captures the consensus over which objects should be mentioned. $\rho^I_k$ will be small for words/object labels that can are mentioned across most references (using the exact word or a semantically similar word), and large for those that are mentioned only by a few. 

The proposed approach of integrating object importance replaces the cost $c(i,j)$ in Eq.~\ref{eq:WMD} with a weighted cost $c'(i, j | R^I)$ to move from word $i$ to word $j$ given references $R^I$:

\begin{equation}
c^{\prime}(i,j | R^I) = \|\rho^I_ix_i - \rho^I_jx_j\|_2^p
\label{eq:newcost}
\end{equation}

\begin{figure*}[t]
\centering
\includegraphics[width=0.7\linewidth]{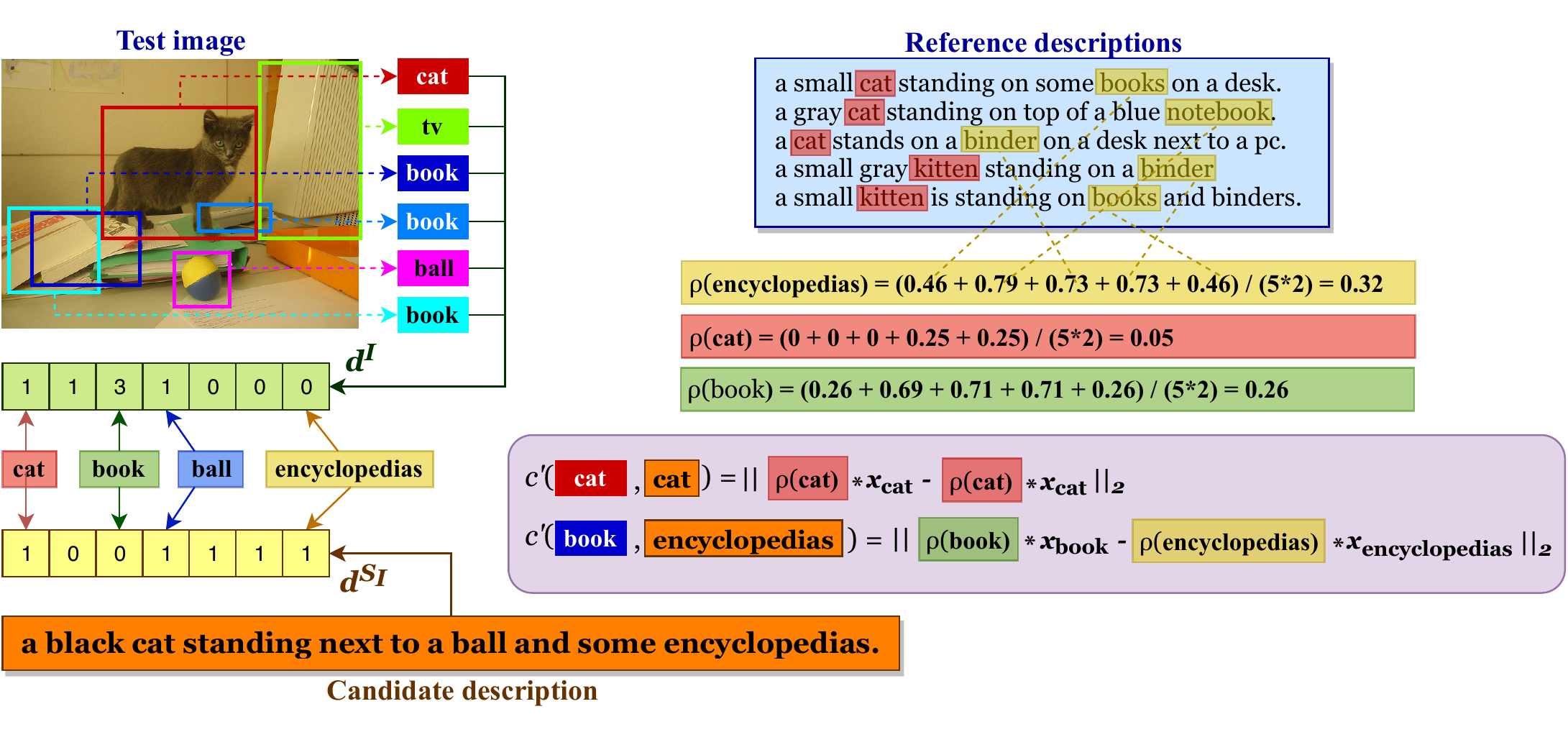}
\caption{Illustration of how object importance based on human references is computed and integrated into \vifidel. In this example, we compute the weights $\rho^I_k$ for \textit{encyclopedias}, \textit{cat} and \textit{books}. The most similar word to \textit{encyclopedias} in each reference description include \textit{books}, \textit{notebook} and \textit{binder} according to the cosine distance between their word embeddings. These are averaged to obtain a consensus penalty score. The word \textit{cat} has a low penalty score because it has either an exact match or a very close semantic match (\textit{kitten}) to all references. The penalty scores are then used as weights to compute the cost $c'(book, encyclopedias | I)$ between the object label \textit{book} from the image and the word \textit{encyclopedias} in the candidate description.}
\label{fig:importance}
\end{figure*}

\noindent The updated Eq.~\ref{eq:WMD} is then used in Eq.~\ref{eq:vifidel} to compute a \vifidel score weighted by object importance. Figure~\ref{fig:importance} illustrates a concrete example of \vifidel's object importance model using human references, showing how the cost $c'(i, j | R^I)$ is calculated.

\section{Experiments}
\label{sec:evaluation}

We experiment with \vifidel in two datasets: the PASCAL-50S Consensus dataset ($\S$\ref{sec:pascal50s}) and human ratings on MSCOCO ($\S$\ref{sec:composite}). We compare \vifidel against commonly used IDG metrics. 

\subsection{Visual annotation and detectors}
\label{sec:visual}

We test the performance of the following metric variants, where {\bf \vifidel$_{gold}$} and {\bf \vifidel$_{D500}$} and their union are reported in the main experiments ($\S$\ref{sec:pascal50s} and $\S$\ref{sec:composite}), while the remaining are used for an ablation study ($\S$\ref{sec:discussion}):
\begin{itemize}
\itemsep0em 
\item \textbf{\vifidel$_{gold}$}: This variant uses gold standard, object-level annotations provided by the respective datasets. For the PASCAL-50S dataset, we use the annotations for 20 pre-defined object categories (\textit{person}, \textit{car}, \textit{cow}, etc.) provided by the PASCAL VOC challenge~\cite{EveringhamEtAl:2015}. For MSCOCO, we use annotations for 80 object categories provided by MSCOCO~\cite{LinEtAl:2014}. In both datasets, the reference descriptions are sourced independent of the image annotations; thus there is no direct correspondence between the visual annotations and the descriptions.
\item \textbf{\vifidel$_{D80}$}: This variant uses the output of an object detector pre-trained on the MSCOCO dataset, for 80 MSCOCO categories. We use the TensorFlow Object Detection API~\cite{HuangEtAl:2017} for this purpose\footnote{\texttt{faster\textunderscore rcnn\textunderscore inception\textunderscore resnet\textunderscore v2\textunderscore atrous\textunderscore\\ coco} from \url{https://github.com/tensorflow/models/blob/master/research/object_detection/g3doc/detection_model_zoo.md}.}. We set 0.6 as the confidence threshold for detected objects.
\item \textbf{\vifidel$_{D500}$}: This variant uses the output of an object detector, pre-trained on the Open Images dataset~\cite{KrasinEtAl:2017} with bounding box annotations for 545 object categories. Again, we use the TensorFlow Object Detection API\footnote{ 
\texttt{faster\textunderscore rcnn\textunderscore inception\textunderscore resnet\textunderscore v2\textunderscore atrous\textunderscore \\ oid}.}, and set the confidence threshold to 0.4.
\item \textbf{\vifidel$_{gold\,\cup\,D500}$}: We combine the outputs of the $gold$ annotation and $D500$ detector and use unique object labels from the combination.
\item \textbf{\vifidel$_{D80\,\cup\,D500}$}: We combine the outputs of the $D80$ and $D500$ detectors.
\end{itemize}

In this paper, we use only the output labels of the detectors, and represent the content of an image $I$ as a vector of normalised frequencies over object labels, $d^I$. A discussion on how the performance of the metric can vary according to the quality of the objects available is given in $S$\ref{sec:discussion}. The ideal setting would count on a comprehensive list of objects given by humans. 




\subsection{Accuracy on PASCAL-50S}
\label{sec:pascal50s}

In this section, we focus on the PASCAL-50S dataset and tackle the binary forced-choice task of predicting: \textit{``which description is more similar to A: B or C?''}, as proposed by \citet{VedantamEtAl:2015}. We focus on the variant comparing two machine generated captions. The dataset contains multiple crowdsourced image description for each of 1,000 images from the UIUC PASCAL dataset~\cite{RashtchianEtAl:2010}.

Evaluation of system outputs as relative rankings has long been established as the best practice in many fields where language outputs are produced and no single correct output exists. The WMT yearly evaluation campaigns for machine translation \cite{BojarEtAl:2016}, for example, have argued that relative ranking leads to more reliable judgments than absolute scores.  We therefore consider our findings on this dataset as the most important.  

For the binary forced-choice task, \citet{VedantamEtAl:2015} collected 48 descriptions \textit{A} per image, and formed pairs of descriptions \textit{B} and \textit{C} from  machine generated descriptions and/or the remaining two human descriptions. This corresponds to the ({\bf MM}) setting of the dataset: comparing two machine generated descriptions.
Arguably, this is most interesting subtask from a practical point of view, which is the setting in which evaluation is generally performed. 

The dataset thus consists of 1,000 \textit{(B,C)} pairs. 
%
Crowd-sourced gold standard annotations were provided for each 48$\times$1000 \textit{(A,B,C)} triplet, and the `consensus' binary label per \textit{(B,C)} pair was obtained by majority vote across 48 references. 
 We also provide results on the other splits in the Appendix~\ref{sec:fullpascal50s}.

\begin{table}[t]
\centering
\resizebox{0.8\linewidth}{!}{ 
\begin{tabular}{lccll}
\toprule
& \multicolumn{4}{c}{References} \\
& 0 & 1 & 5 & 48 \\
\midrule
\textbf{BLEU$_1$} & - & 0.58 & 0.61 & 0.62\\
\textbf{BLEU$_4$} & - & 0.56 & 0.59 & 0.60\\
\textbf{ROUGE$_L$} &  - & 0.58 & 0.61 & 0.64\\
\textbf{METEOR} & - & 0.62 & 0.66 & 0.68\\
\textbf{CIDEr} & - & 0.60 & 0.66 & 0.68\\
\textbf{SPICE} & - & 0.65 & 0.69 & 0.69\\ 
\textbf{WMD}$_{\text{best}}$ & - & 0.66 & 0.70 & 0.70\\ 
\textbf{WMD}$_{\text{worst}}$ & - & 0.66 & 0.66 & 0.66\\ 
\textbf{LM} & 0.54 & 0.54 & 0.54 & 0.54\\ 
\midrule
\textbf{\vifidel$_{gold}$} & 0.68 & 0.69 & 0.70 & {\bf 0.71} \\
\textbf{\vifidel$_{D500}$} & {\bf 0.69} & {\bf 0.71} & \textbf{0.72} &  {\bf 0.71}\\
\midrule
\textbf{\vifidel+LM} & \textbf{0.69} & 0.70 & 0.71 & 0.71 \\
\textbf{\vifidel+CIDEr} & \textbf{0.69} & \textbf{0.71} & \textbf{0.72} & {\bf 0.72} \\
\bottomrule
\end{tabular}
}
\caption{Accuracy of \vifidel on the PASCAL-50S binary forced-choice task using 0, 1, 5 and 48 references,  comparing two machine generated descriptions.} 
\label{tbl:pascal50s1}
\end{table}

\begin{figure}[t]
\centering
\includegraphics[width=\linewidth]{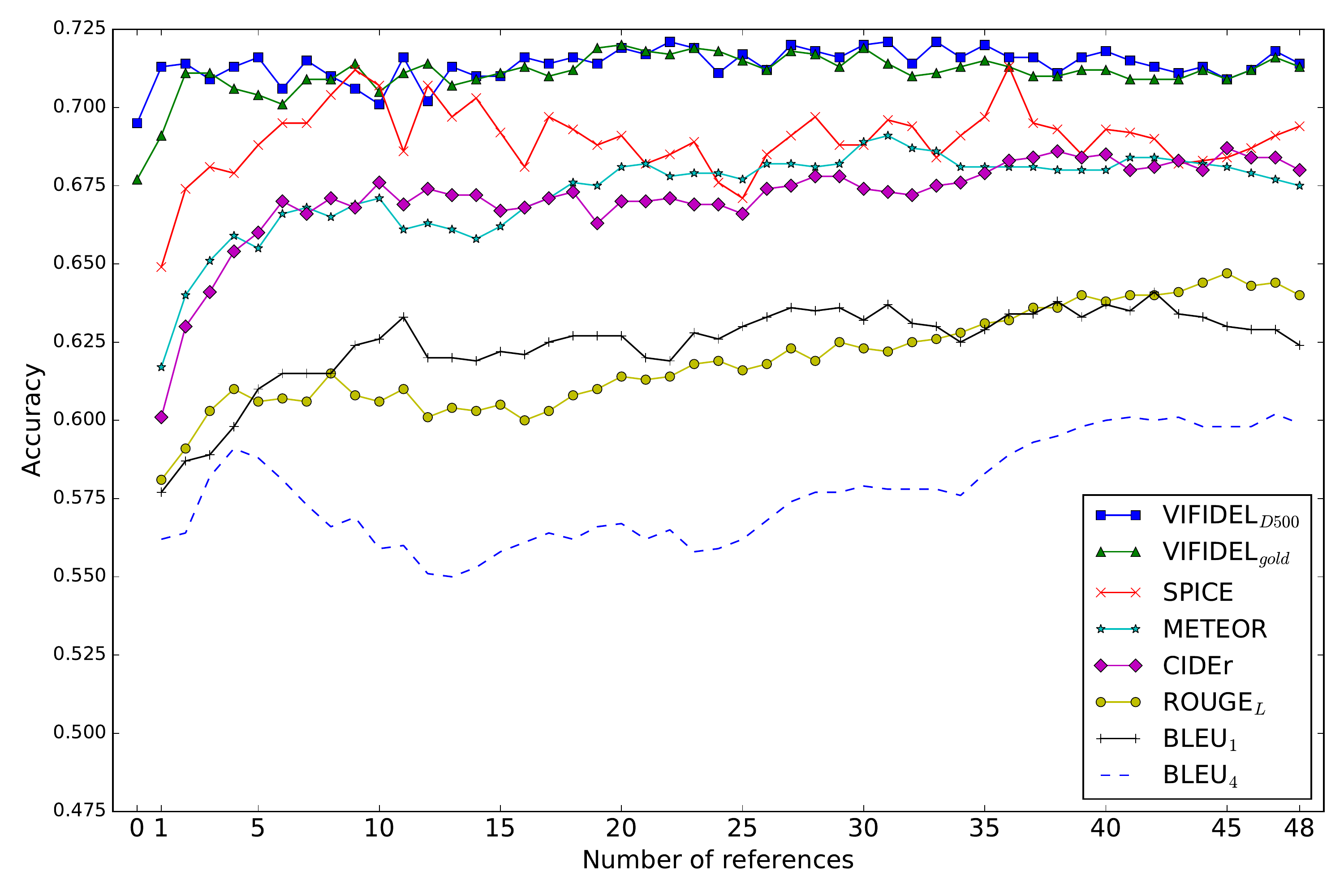}
\caption{Accuracy of consensus predictions on PASCAL-50S, across increasing number of references, for the Machine-Machine (MM) subset.} 
\label{fig:accuracy}
\end{figure}

\paragraph{Results and discussion}
Table~\ref{tbl:pascal50s1} presents the accuracies for the binary forced-choice task for \vifidel, compared to the most commonly used IDG metrics: BLEU$_1$, BLEU$_4$, ROUGE$_L$, METEOR, CIDEr-D, and SPICE\footnote{We use the official COCO (\url{https://github.com/tylin/coco-caption}) and SPICE (\url{https://github.com/peteanderson80/SPICE}) scripts.}. We also report results with the standard WMD as the metric and an RNN language model trained on training captions from MSCOCO~\cite{ChenEtAl:2015}. Note that standard WMD is already a strong metric, but it relies on references, like all other metrics in the table and different from \vifidel. \vifidel only uses references to up-weight or down-weight the matches between image objects and words in the captions. 
Notable is the metric's performance based {\em solely} on image information: \vifidel can distinguish between a correct and incorrect description -- the difficult task of differentiating between two machine generated descriptions, agreeing with human judgments more often than any other IDG metric that uses a reference. With one reference description, \vifidel gave the highest accuracy, 
suggesting that image-side information is indeed helpful for evaluating the quality of image descriptions when comparing two machine generated descriptions. 

Even when we take more references into account, \vifidel is as good or better than reference-based metric. With 5 references (a feasible number from a practical perspective), \vifidel  achieves the highest accuracy with a slight improvement in score over zero or one reference. Thus, the object weighting scheme can be seen to help \vifidel focus on which objects are more important. With 48 references (the maximum possible, only available in PASCAL-50S), \vifidel is still more accurate than existing metrics, showing that visual fidelity is an important factor for rating machine generated descriptions.

Figure~\ref{fig:accuracy} depicts accuracies over an increasing number of reference descriptions, starting from zero references (only defined for \vifidel). 
We note that \vifidel is more stable and consistently \emph{outperforms} other metrics for all numbers of references. 
\vifidel$_{D500}$ has a very slight advantage over \vifidel$_{gold}$, 
most likely because the visual information is richer in the former. 

Overall, we conclude that measuring visual fidelity is important especially for ranking two machine generated descriptions, arguably the most important evaluation setting, and that \vifidel can measure visual fidelity by explicitly using information derived from images, particularly when few or no references are available. 

\subsection{Correlation with human judges}
\label{sec:composite}

\begin{table*}[!t]
\centering
\resizebox{0.85\linewidth}{!}{ 
\begin{tabular}{lcc @{\hspace{1.2cm}} cc @{\hspace{1.2cm}} cc}
\toprule
& \multicolumn{2}{c}{5 Refs} &  \multicolumn{2}{c}{1 Ref} & \multicolumn{2}{c}{0 Ref}\\
\noalign{\smallskip}
\cline{2-3}\cline{4-7}
\noalign{\smallskip}
Metric & \textbf{Relevance} & \textbf{Thoroughness} & \textbf{Relevance} & \textbf{Thoroughness} & \textbf{Relevance} & \textbf{Thoroughness}\\
\midrule
\textbf{BLEU$_1$} & 0.26 & 0.25 & 0.23 & 0.20 & - & - \\
\textbf{BLEU$_4$} & 0.26 & 0.24 & 0.22 & 0.20 & - & - \\
\textbf{ROUGE$_L$} & 0.28 & 0.26 & 0.24 & 0.22 & - & - \\
\textbf{METEOR} & 0.30 & 0.27 & 0.25 & 0.22 & - & - \\
\textbf{CIDEr} & 0.32 & 0.28 & 0.26 & 0.24 & - & - \\
\textbf{SPICE} & \textbf{0.35} & \textbf{0.31} & 0.27 & \textbf{0.27} & - & - \\
\textbf{WMD$_{\text{Best}}$} & 0.30 & 0.26 & 0.28 & 0.26& - & -  \\
\textbf{WMD$_{\text{Worst}}$} & 0.29 & 0.26 & 0.28 & 0.25& - & -  \\
\textbf{LM} & -0.10 & -0.16 & -0.10 & -0.16 & -0.10 & -0.16\\
\midrule
\textbf{\vifidel$_{\text{no refs},{gold\,\cup\,D500}}$} & 0.22 & 0.21 & 0.22 & 0.21 & 0.21 & 0.22 \\
\textbf{\vifidel$_{gold\,\cup\,D500}$} & 0.30 & 0.27 & \textbf{0.29} & \textbf{0.27} & - & -\\
\midrule
\textbf{VIFIDEL$_{\text{no refs}}+$LM} & 0.21 & 0.21 & 0.21 & 0.21 & 0.21 & 0.21\\
\textbf{VIFIDEL$+$CIDEr} & 0.29& 0.29 & \textbf{0.29} & \textbf{0.28} & - & - \\
\bottomrule
\end{tabular}
}
\caption{Spearman's rank correlation coefficient between automatic metrics and human judgment scores over the MSCOCO portion of COMPOSITE dataset.}
\label{tab:idg}
\end{table*}

We measure correlation with human judgments on the MSCOCO portion of the COMPOSITE dataset~\cite{AdityaEtAl:2018}. The dataset contains 
$2,007$ candidate images from the MSCOCO dataset~\cite{LinEtAl:2014} with annotations from AMT workers. These annotations consist of judgments on a scale of 1 (low) to 5 (high) regarding (i) relevance of a description to the image and (ii) thoroughness of a description given an image. Candidate descriptions were sampled from human references and image description systems. To make our settings closer to those of real system evaluations, we consider only system generated descriptions as candidates  and evaluate the Spearman's correlation between the automated metrics and human judgments. As mentioned in $\S$\ref{sec:pascal50s}, absolute human judgments, especially for subjective tasks such as this, can be very subjective and therefore less reliable, especially when a single judgment is collected per description.

\paragraph{Results and discussion}
We summarise our results in Table~\ref{tab:idg}. For these experiments, we report ${gold\,\cup\,D500}$ detectors for object information, as this was the best overall performance (non-gold and other variants can be seen in $\S$\ref{sec:discussion}). Our key finding is that \vifidel using no reference descriptions obtains comparable (albeit still lower) correlation to metrics like BLEU and ROUGE with only one reference description. 
The gap between \vifidel and such metrics is the inability of the former to capture fluency, since it relies essentially on bag of word embeddings.

\subsection{Combining \vifidel with other metrics} 
As mentioned, \vifidel is a fine-grained metric that exclusively evaluates the visual fidelity of an image description. As such, by design, an IDG system may achieve high \vifidel scores simply by listing all objects depicted in the image. We do not consider this an issue, since we are in effect evaluating \emph{what} the image description describes (visual fidelity), rather than \emph{how} the image is being described (fluency). The latter can instead be evaluated separately with a metric designed specifically for the purpose. Thus, a system that simply lists all objects will result in a high \vifidel score but a low fluency score. This is in line with our vision of evaluating the specific capabilities of IDG systems to clearly understand \emph{how} one system is better than another.
While this is not the aim of the paper, we also demonstrate how \vifidel can also be combined with  fluency-based metrics to evaluate image descriptions as a single conflated metric. More specifically, we explore combining VIFIDEL with two different fluency-based strategies: an RNN language model and CIDEr (other metrics are possible, CIDEr provides a good compromise between performance and efficiency). In both cases, we simply averaged the scores of the two metrics. The results are shown at the bottom of Tables~\ref{tbl:pascal50s1} and~\ref{tab:idg}. On average, the addition of the fluency-based metric is complementary. {\vifidel$+$CIDEr} is better performing than {\vifidel$+$LM}. This is expected as LM only provides perplexity scores given a description, while CIDEr explicitly measures quality against references.  We note that better combinations can potentially be achieved by learning weights for a weighted average and optimising the training towards fluency.  

\begin{table}[ht!]
\centering
\resizebox{\linewidth}{!}{ 
\begin{tabular}{lcc @{\hspace{1.2cm}} cc}
\toprule
& \multicolumn{2}{c}{Without Frequency} &  \multicolumn{2}{c}{With Frequency}\\
\noalign{\smallskip}
\cline{2-5}
\noalign{\smallskip}
Object Detectors & \textbf{Relevance} & \textbf{Thoroughness} & \textbf{Relevance} & \textbf{Thoroughness} \\
\midrule
\textbf{$gold$} & 0.25 & 0.24 & 0.25 & 0.23\\
\textbf{$D80$} & 0.24 & 0.23 & 0.24 & 0.22\\
\textbf{$D500$} & 0.25 & 0.24 & 0.25 & 0.23 \\
\textbf{${gold\,\cup\,D500}$} & \textbf{0.29} & \textbf{0.27} & \textbf{0.28} & \textbf{0.26}\\
\textbf{${D80\,\cup\,D500}$} & 0.28 & 0.26 & 0.27 & \textbf{0.26}\\
\bottomrule
\end{tabular}
}
\caption{Ablation experiments on \vifidel 
on the MSCOCO part of the COMPOSITE dataset with different object detectors and with vs.\ without frequency counts for objects.}
\label{tab:ablate}
\end{table}

\subsection{Ablation studies}
\label{sec:discussion}

\paragraph{Effect of object detectors and frequency counts:}
Here we study the effect of using different object detectors. 
We also investigate the contribution of frequency counts in $d^I$, by binarising the frequency counts to indicate only the presence and absence of the objects. The hypothesis is that the number of object instances may be useful for evaluating visual fidelity.
The results are summarised in Table~\ref{tab:ablate}. The combination of $gold$ and $D500$ object detectors performed the best for the dataset. The $gold$ object information is only slightly better than $D80$ prediction-based object information. Interestingly, $D500$, which is more fine-grained, performs as well as the $80$-category $gold$ object information.  
Using a binarised $d^I$ seemed to give comparable correlation, perhaps even with a marginal edge over its frequency-based counterpart. 
We postulate that this could be because frequency counts are likely to be mentioned in the descriptions via quantifiers and other morphological and typological variants, which cannot be easily mapped to the frequency of detected objects. 

\begin{figure}[t]
\centering
\resizebox{0.95\linewidth}{!}{
{ \tiny
\begin{tabular}{p{4cm} p{2.5cm} p{0.5cm} }
\toprule
\multirow{4}
{*}{\includegraphics[height=2.8cm]{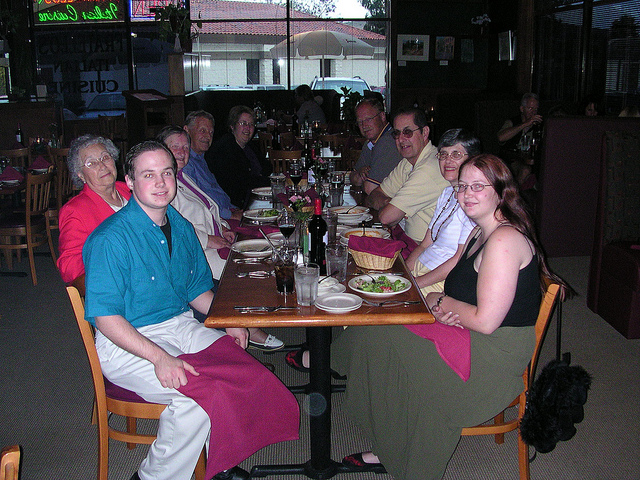}} & person & 0.75  \\ \cmidrule{2-3}
& person, dining-table & 0.83 \\ \cmidrule{2-3}
& \multicolumn{1}{p{2.5cm}}{\raggedright  person, dining-table, umbrella, handbag, bottle} & 0.74 \\\cmidrule{2-3}
& \multicolumn{1}{p{2.5cm}}{\raggedright person, car, backpack, umbrella, handbag, bottle, dining-table, cup, fork, knife} & 0.73\\\cmidrule{2-3}
& \multicolumn{1}{p{2.5cm}}{\raggedright person, car, backpack, umbrella, handbag, bottle, wine-glass, cup, fork, knife, spoon, bowl, broccoli, chair, dining-table} & 0.70\\ 

\bottomrule
\end{tabular}
}
}
\caption{\vifidel score changes with different numbers of object annotations.}
\label{fig:ablationeg}
\end{figure}

\paragraph{Effect of number of detected objects:}
We used a pre-trained captioning system~\cite{AndersonEtAl:2018} to generate captions on a sample of images from the MSCOCO validation set that have different gold object annotations. We present two examples in Table~\ref{tab:ablateobj}. 
\begin{table}[t!]
\centering
\resizebox{0.95\linewidth}{!}{ 
\begin{tabular}{lcll}
\toprule
\vifidel & Objects & Caption & References\\
\midrule
$0.79$ & \multicolumn{1}{p{2cm}}{truck} & \multicolumn{1}{p{4cm}}{\raggedright a small truck sitting on top of a field} & \multicolumn{1}{p{6cm}}{\raggedright 1. A orange tractor sitting on top of a lush green field.\newline
2. A snowplow truck with two snowplows on it \newline
3. Farm equipment truck in a field near a road. \newline
4. an image of  a truck with the scaffolding \newline
5. The yellow earth mover sits in the field in front of the pole.}\\
\midrule
$0.70$ & \multicolumn{1}{p{2cm}}{\raggedright person, car, backpack, umbrella, handbag, bottle, wine-glass, cup fork, knife, spoon, bowl, broccoli, chair, dining-table} & \multicolumn{1}{p{4cm}}{\raggedright a table full of people at the restaurant} & \multicolumn{1}{p{6cm}}{\raggedright 1. A group of people sitting around a wooden table with food.\newline
2. The nine people smile as they sit at a dinner table.\newline
3. A table full of people that are eating at a restaurant.\newline
4. Co-workers often get together after a long day at work.\newline
5. A group of people that are sitting around a table.}\\
\bottomrule
\end{tabular}
}
\caption{Ablation study on the number of detected objects:   one detected object and fifteen detected objects.}
\label{tab:ablateobj}
\end{table}
%
In the first example the image contains only one object. Here \vifidel relies both on the object and on the semantic similarity between the caption and references. In the second example there are fifteen objects, however some are more important than others for describing the images. \vifidel gives higher importance to objects that are mentioned in the references. 
In Figure~\ref{fig:ablationeg} we further explore the third example from our ablation studies by computing \vifidel for different subsets of object annotations, ranging from one object to fifteen object annotations. We see that, as the number of objects increase from one to two, the \vifidel score also increases, given that these objects are also mentioned in the system caption. However, with more objects the scores go down, since these are not mentioned in the caption, but the decrease is gradual, even with 15 objects. This happens because such additional objects do not seem too relevant to humans, as these are mostly not mentioned in the references. Therefore, the scores from \vifidel change with both number of annotations and the reference descriptions, while metrics like METEOR ($0.354$) and SPICE ($0.320$) would remain constant.

\paragraph{Effect of word representations:} 
We also studied the effect of various pre-trained embeddings and found that the pre-trained model of \emph{word2vec}  $300$-dimensional \emph{CBOW} embeddings~\cite{MikolovEtAl:2013} is slightly better than \emph{GLoVe} \cite{PenningtonEtAl:2014} and \emph{FastText} embeddings~\cite{JoulinEtAl:2016}.
This could be because of the amount of data on which these were trained. \emph{FastText} embeddings had similar performance as word2vec  embeddings even when only trained on the Wikipedia as corpus. For consistency, we used the \emph{word2vec} embeddings pre-trained on Google News.

\section{Conclusions}

We have introduced a new metric for image description evaluation that goes beyond comparing descriptions to human references and is explicitly based on object-level image information. Our  hypothesis is that the use of image information provides a more reliable pathway for measuring the fidelity of a description for a given image. Further, the metric relies on off-the shelf object detectors and word-embeddings and computes the scores in a semantic space. Our analysis on two of the most widely used datasets for metric comparison shows that our metric correlates well with human judgments, and is particularly well suited when few or no reference description is available. The metric  performs comparatively for gold and predicted annotations on objects and is lightweight in terms of dependency on linguistic resources.
Our implementation of \vifidel can be accessed from: \url{https://github.com/ImperialNLP/vifidel}

%

\section*{Acknowledgements}

The authors thank the anonymous reviewers and area chairs for giving up their time to provide useful feedback on an earlier draft of this paper. This work was supported by the MultiMT project (H2020 ERC Starting Grant No. 678017). This work was also supported by the MMVC project, via an Institutional Links grant, ID 352343575, under the Newton-Katip Celebi Fund partnership. The grant is funded by the UK Department of Business, Energy and Industrial Strategy (BEIS) and Scientific and Technological Research Council of Turkey (T{\"U}B{\.I}TAK) and delivered by the British Council. For further information, please visit \url{http://www.newtonfund.ac.uk}.

\bibliography{refs}
\bibliographystyle{acl_natbib}

\appendix

\section{Accuracy on PASCAL-50S (all groups)}
\label{sec:fullpascal50s}
The original PASCAL-50S dataset consisted of four groups: (i) \textbf{HC} (Human-Human correct: both correct human descriptions); (ii) \textbf{HI} (Human-Human incorrect: one correct human description and one human description from another random image); (iii) \textbf{HM} (Human-Machine: one correct human description and one machine generated description); (iv) \textbf{MM} (Machine-Machine: both machine generated descriptions). 
We summarise the accuracy of \vifidel and various evaluation metrics on the PASCAL-50S binary forced-choice task in Tables~\ref{tbl:pascal50s1full} and \ref{tbl:pascal50s2full}.

\begin{table}[h]
\begin{center}
\resizebox{\linewidth}{!}{ 
\begin{tabular}{lccccc @{\hspace{1.2cm}} ccccc}
\toprule
\multirow{2}{*}{Metric} & \multicolumn{5}{c}{No references} & \multicolumn{5}{c}{1 reference} \\
\noalign{\smallskip}
\cline{2-11}
\noalign{\smallskip}
& HC & HI & HM & MM & \textbf{all} & HC & HI & HM & MM & \textbf{all}\\
\midrule
\textbf{BLEU$_1$} & - & - & - & - & - & 0.61 & 0.86 & 0.88 & 0.58 & 0.73 \\
\textbf{BLEU$_4$} & - & - & - & - & - & 0.59 & 0.83 & 0.80 & 0.56 & 0.70 \\
\textbf{ROUGE$_L$} & - & - & - & - & - & 0.61 & 0.87 & 0.86 & 0.58 & 0.73 \\
\textbf{METEOR} & - & - & - & - & - & 0.61 & 0.91 & {\bf 0.89} & 0.62 & {\bf 0.76} \\
\textbf{CIDEr} & - & - & - & - & - & {\bf 0.62} & 0.91 & 0.85 & 0.60 & 0.75 \\
\textbf{SPICE} & - & - & - & - & - & 0.58 & 0.87 & 0.77 & 0.65 & 0.72 \\ 
\midrule
\textbf{\vifidel$_{gold}$} & {\bf 0.57} & 0.91 & 0.53 & 0.68 & 0.67 & {\bf 0.62} & {\bf 0.96} & 0.65 & 0.69 & 0.73 \\
\textbf{\vifidel$_{D500}$} & {\bf 0.57} & {\bf 0.93} & {\bf 0.64} & {\bf 0.69} & {\bf 0.71} & 0.59 & 0.95 & 0.74 & {\bf 0.71} & 0.75 \\
\bottomrule
\end{tabular}
}
\end{center}
\caption{Accuracy of \vifidel  and other IDG metrics on the PASCAL-50S binary forced-choice task using no references and one reference.}
\label{tbl:pascal50s1full}
\end{table}

\begin{table}[h]
\begin{center}
\resizebox{\linewidth}{!}{ 
\begin{tabular}{lccccc @{\hspace{1.2cm}} ccccc}
\toprule
\multirow{2}{*}{Metric} & \multicolumn{5}{c}{5 references} & \multicolumn{5}{c}{48 references} \\
\noalign{\smallskip}
\cline{2-11}
\noalign{\smallskip}
& HC & HI & HM & MM & \textbf{all} & HC & HI & HM & MM & \textbf{all}\\
\midrule
\textbf{BLEU$_1$} & 0.64 & 0.95 & 0.91 & 0.61 & 0.78 & 0.63 & 0.98 & 0.94 & 0.62 & 0.79 \\
\textbf{BLEU$_4$} & 0.62 & 0.93 & 0.85 & 0.59 & 0.75 & 0.64 & 0.97 & 0.92 & 0.60 & 0.78 \\
\textbf{ROUGE$_L$} & 0.66 & 0.95 & 0.93 & 0.61 & 0.78 & 0.67 & 0.98 & 0.95 & 0.64 & 0.81 \\
\textbf{METEOR} & 0.66 & \textbf{0.98} & \textbf{0.94} & 0.66 & \textbf{0.81} & 0.65 & \textbf{0.99} & \textbf{0.96} & 0.68 & \textbf{0.82} \\
\textbf{CIDEr} & \textbf{0.67} & \textbf{0.98} & 0.90 & 0.66 & 0.80 & \textbf{0.69} & \textbf{0.99} & 0.92 & 0.68 & \textbf{0.82} \\
\textbf{SPICE} & \textbf{0.67} & 0.97 & 0.89 & 0.69 & 0.80 & 0.62 & \textbf{0.99} & 0.94 & 0.69 & 0.81 \\
\midrule
\textbf{\vifidel$_{gold}$} & 0.65 & \textbf{0.98} & 0.66 & 0.70 & 0.75 & 0.64 & 0.98 & 0.67 & \textbf{0.71} & 0.75 \\
\textbf{\vifidel$_{D500}$} & 0.64 & 0.97 & 0.75 & \textbf{0.72} & 0.77 & 0.63 & 0.97 & 0.76 & \textbf{0.71} & 0.77\\
\bottomrule
\end{tabular}
}
\end{center}
\caption{Accuracy of \vifidel and other IDG metrics on the PASCAL-50S binary forced-choice task using 5 
and 48 references. 
}\label{tbl:pascal50s2full}
\end{table}

\end{document}